\documentclass[10pt,twocolumn,letterpaper]{article}
\usepackage{cvpr}
\usepackage{booktabs}
\usepackage{multirow}
\usepackage{adjustbox}
\usepackage{tabularx, colortbl}
\usepackage{bbding}
\usepackage{url}            
\usepackage{booktabs}       
\usepackage{amsfonts}       
\usepackage{nicefrac}       
\usepackage{microtype}      
\usepackage{xcolor}         
\usepackage{amsmath}
\usepackage{amssymb}
\usepackage{enumitem}
\usepackage{amsthm}

\newtheorem{theorem}{Theorem}

\newtheorem{assumption}{Assumption}

\DeclareMathOperator{\x}{\mathbf{x}}
\DeclareMathOperator{\Phix}{\mathbf{\Phi x}}
\DeclareMathOperator{\y}{\mathbf{y}}
\DeclareMathOperator{\PhiTy}{\mathbf{\Phi}^T \mathbf{y}}
\DeclareMathOperator{\PhiTu}{\mathbf{\Phi}^T \mathbf{u}}
\DeclareMathOperator{\bfPhi}{\mathbf{\Phi}}

\DeclareMathOperator{\R}{\mathbb{R}}
\DeclareMathOperator{\I}{\mathbf{I}}
\DeclareMathOperator{\zz}{\mathbf{z}}
\DeclareMathOperator{\rr}{\mathbf{r}}

\DeclareMathOperator{\bfs}{\mathbf{s}}
\DeclareMathOperator{\hh}{\mathbf{h}}
\DeclareMathOperator{\bfu}{\mathbf{u}}
\DeclareMathOperator{\barAlpha}{\bar{\alpha}}
\DeclareMathOperator{\bfd}{\mathbf{d}}

\usepackage[accsupp]{axessibility}  

\definecolor{cvprblue}{rgb}{0.21,0.49,0.74}
\usepackage[pagebackref,breaklinks,colorlinks,allcolors=cvprblue]{hyperref}

\def\paperID{3637}
\def\confName{CVPR}
\def\confYear{2025}

\title{Using Powerful Prior Knowledge of Diffusion Model in Deep Unfolding Networks for Image Compressive Sensing}

\author{Chen Liao, Yan Shen\textsuperscript{*}, Dan Li, Zhongli Wang\\
	Beijing Jiaotong University, China\\
	{\tt\small \{liaochen, sheny, lidan102628, zlwang\}@bjtu.edu.cn}
}

\begin{document}
\maketitle

\begin{abstract}
	Recently, Deep Unfolding Networks (DUNs) have achieved impressive reconstruction quality in the field of image Compressive Sensing (CS) by unfolding iterative optimization algorithms into neural networks. The reconstruction quality of DUNs depends on the learned prior knowledge, so introducing stronger prior knowledge can further improve reconstruction quality. On the other hand, pre-trained diffusion models contain powerful prior knowledge and have a solid theoretical foundation and strong scalability, but it requires a large number of iterative steps to achieve reconstruction. In this paper, we propose to use the powerful prior knowledge of pre-trained diffusion model in DUNs to achieve high-quality reconstruction with less steps for image CS. Specifically, we first design an iterative optimization algorithm named Diffusion Message Passing (DMP), which embeds a pre-trained diffusion model into each iteration process of DMP. Then, we deeply unfold the DMP algorithm into a neural network named DMP-DUN. The proposed DMP-DUN can use lightweight neural networks to achieve mapping from measurement data to the intermediate steps of the reverse diffusion process and directly approximate the divergence of the diffusion model, thereby further improving reconstruction efficiency. Extensive experiments show that our proposed DMP-DUN achieves state-of-the-art performance and requires at least only 2 steps to reconstruct the image. Codes are available at \url{https://github.com/FengodChen/DMP-DUN-CVPR2025}.
\end{abstract}

\begin{figure}[t]
	\centering
	\includegraphics[width=\linewidth]{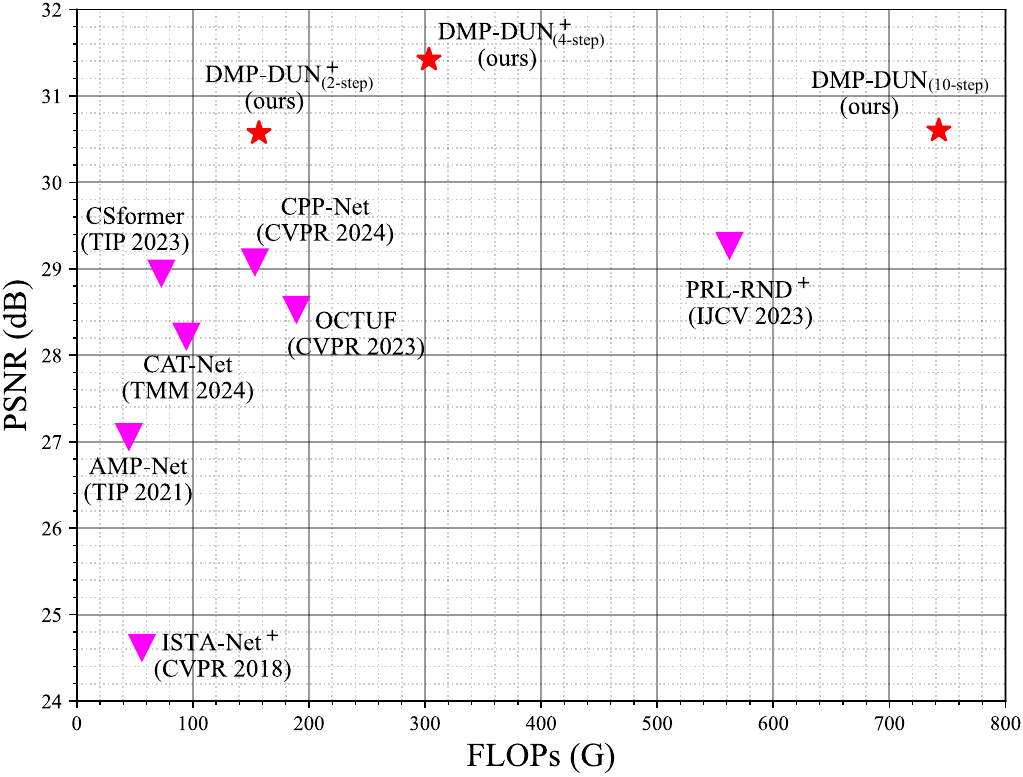}
	\caption{Comparison of the average PSNR and FLOPs of our proposed DMP-DUN with other methods on Urban100\cite{Dong2018DenoisingPD}, under various CS ratio of 1\%, 4\%, 10\%, 25\% and 50\%.}
	\label{fig:psnr_flops_compare}
\end{figure}
\begin{figure}[t]
	\centering
	\includegraphics[width=\linewidth]{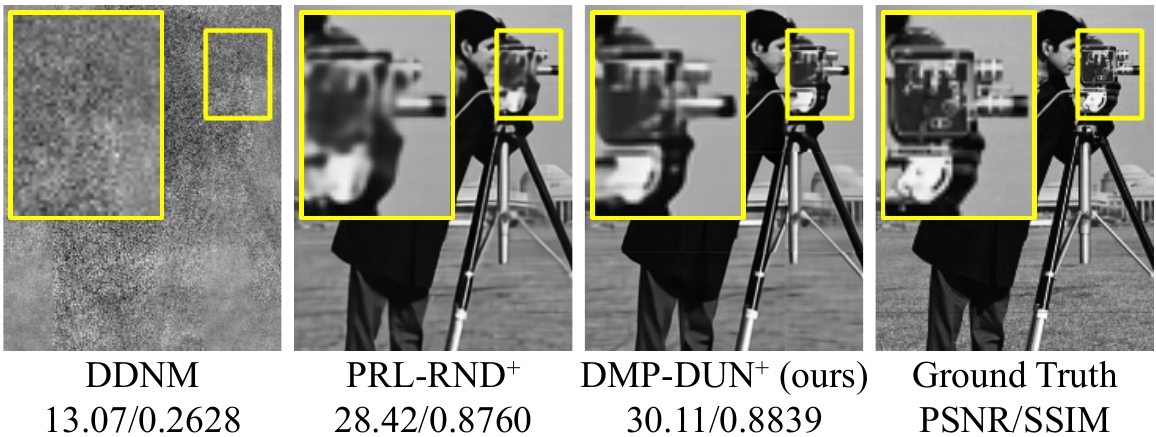}
	\caption{The reconstruction effects of different methods using only 2 steps under CS ratio = 10\%.}
	\label{fig:pre_compare}
\end{figure}

\section{Introduction}
{\def\thefootnote{*}\footnotetext{Corresponding author (Yan Shen: sheny@bjtu.edu.cn)}}

For the reason of utilizing the sparsity of the signal itself or its representation in some transform domain, Compressive Sensing (CS) can reconstruct signal from fewer measurements than conventional methods\cite{candes2006robust, Donoho2006HighDimensionalCS}. Therefore, it has been widely applied in fields such as magnetic resonance imaging (MRI)\cite{Kontogiannis2022PhysicsInformedCS, Lustig2007SparseMT}, snapshot compressive imaging (SCI)\cite{Cheng2021MemoryEfficientNF}, and wireless tele-monitoring\cite{Zhang2012CompressedSF}. Mathematically, the sampling process of CS can be expressed as obtaining CS measurements $\y=\Phix + \epsilon$ from an original signal $\x\in\mathbb{R}^N$ through the linear sensing matrix $\bfPhi\in\mathbb{R}^{M\times N}$ with Gaussian noise $\epsilon \in \mathbb{R}^M$, where $M \ll N$. CS is to infer the orginal signal $\x$ by solving this ill-posed inverse problem with sensing rate $\delta = M / N$. 

To overcome this problem, some transitional iterative optimization algorithms\cite{Daubechies2003AnIT, beck2009fast, Donoho2009MessagepassingAF, Becker2009NESTAAF, afonso2010augmented, DBLP:journals/siamjo/Ochs19} obtain approximate solutions by establishing an optimization model and iteratively optimizing. With the development of deep learning, some methods\cite{kulkarni2016reconnet, shi2017deep, xu2018lapran, shi2019scalable, ye2021csformer, fan2022global} directly use neural networks to fit the mapping from CS measurements to ground truth, thus achieving higher reconstruction speed and quality.

More recently, Deep Unfolding Networks (DUNs) have become novel deep learning methods that are gaining attention in the field of image CS. DUNs\cite{DBLP:conf/icml/GregorL10, yang2018admm, zhang2018ista, zhang2020amp, you2021coast, shen2022transcs, song2021memory, Song2023OptimizationInspiredCT, DBLP:journals/ijcv/ChenSXZ23, guoCPPNet2024, wangUFCNet2024, DBLP:journals/tmm/CuiFZZ24, DBLP:journals/tmm/KongCH24} utilize the strengths of both traditional iterative algorithms and neural networks by unfolding each iteration of an optimization algorithm into a layer of a neural network, hence significantly improving convergence speed and reducing the number of iterations. At the same time, the prior knowledge learned during training is often richer than that manually designed in traditional iterative optimization algorithms, resulting in higher reconstruction quality.

\begin{figure}[t]
	\centering
	\includegraphics[width=\linewidth]{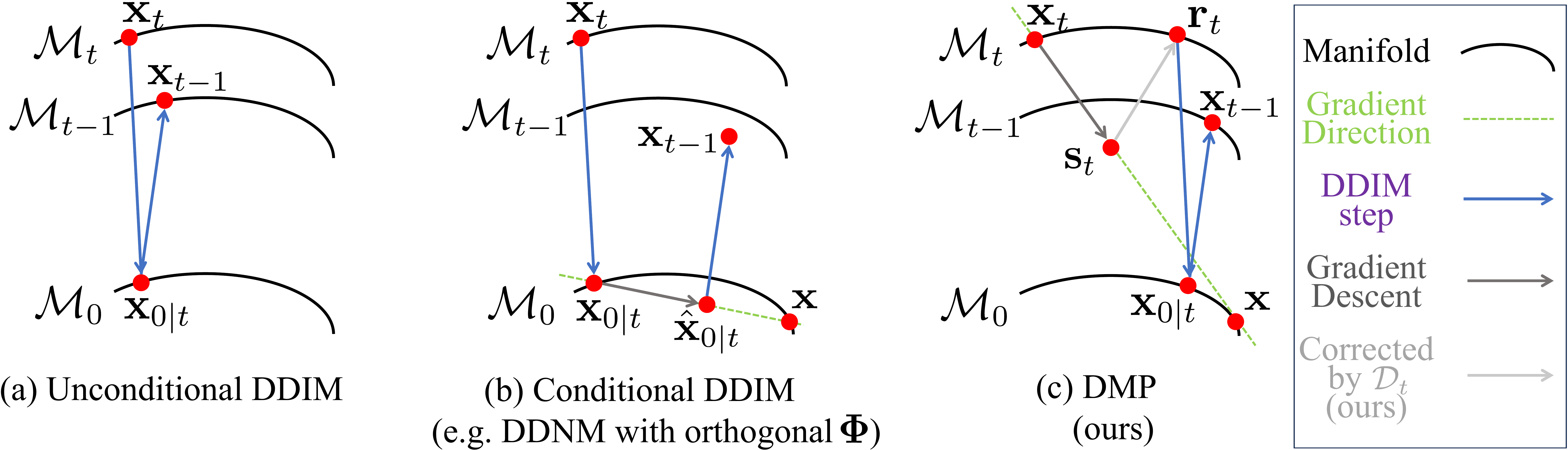}
	\caption{Compared with different type of diffusion models under the perspective of manifold hypothesis.}
	\label{fig:manifold}
\end{figure}

Meanwhile, pre-trained diffusion models\cite{DBLP:conf/nips/HoJA20, DBLP:conf/iclr/SongME21} possess powerful image prior knowledge, enabling high-quality image generation and reconstruction tasks. Currently, many methods\cite{DBLP:conf/nips/KawarEES22, wang2022zero, DBLP:conf/nips/ChungSRY22, he2024manifold, DBLP:conf/iclr/SongVMK23, DBLP:conf/iclr/ChungKMKY23, DBLP:conf/cvpr/FeiLPZYLZ023, DBLP:conf/nips/RoutRDCDS23, DBLP:journals/pami/SahariaHCSFN23} attempt to leverage diffusion models to solve linear inverse problems such as image CS. However, they predominantly require a large number of iterative steps, and perform unsatisfactorily at low CS ratios.

This prompts us to reflect that, on one hand, pre-trained diffusion models possess powerful image prior knowledge but require numerous iterative optimization steps to reconstruct images and have poor performance at low CS ratios. On the other hand, DUNs are a promising approach to mitigating the challenges posed by iterative optimization algorithms, which often necessitate numerous iterations and yield poor performance at low CS ratios. Consequently, can we propose a method that integrates pre-trained diffusion models with DUNs, thereby leveraging the powerful image prior knowledge to accomplish high-quality image CS reconstruction across various CS ratios with a reduced number of iterative steps?

In this paper, we attempt to embed pre-trained diffusion model into DUNs to achieve fast and high-quality image CS reconstruction under various CS ratios.

Firstly, we propose a new iterative optimization algorithm named Diffusion Message Passing (DMP), as shown in \cref{eq:ddm_amp}. This algorithm inspired by the classic CS algorithm Approximate Message Passing (AMP), and can embed one step of the reverse diffusion process into each iteration of DMP. By implementing each single step of reverse diffusion process through a pre-trained diffusion process, the objective of leveraging the powerful prior knowledge of the pre-trained diffusion model in DMP can be realized. Meanwhile, due to the design based on probabilistic graphical ideas and the utilization of statistical properties of data, DMP can provide faster convergence speeds compared to other methods\cite{DBLP:conf/iclr/SongVMK23, DBLP:conf/iclr/ChungKMKY23, DBLP:conf/cvpr/FeiLPZYLZ023, DBLP:conf/nips/RoutRDCDS23, DBLP:journals/pami/SahariaHCSFN23} that use pre-trained diffusion models for reconstruction. Moreover, according to the state evolution of DMP (as shown in \cref{eq:ddm_amp_se}), the manifold-constrained states\cite{he2024manifold} of its variables during reconstruction can be represented as \cref{fig:manifold}(c), which shows that our proposed DMP benefits from its unique state evolution to preserve the manifolds throughout reconstruction process, thereby enhancing the reconstruction quality.

Subsequently, we deeply unfold the proposed DMP into a neural network named DMP-DUN, as shown in \cref{fig:network_summary_detail}. DMP-DUN can directly map CS measurements to the intermediate steps of the reverse diffusion process, and replace the Monte-Carlo SURE method\cite{DBLP:journals/tip/RamaniBU08} with a lightweight trainable network. Furthermore, we obtain the time steps and scaling parameters of DMP-DUN through end-to-end training, hence further improving the reconstruction process. Extensive experiments demonstrate that the reconstruction quality of our proposed method outperforms existing state-of-the-art methods with balanced computing cost  (as shown in \cref{fig:psnr_flops_compare}), and can use only 2 steps of DMP to achieve great reconstruction quality (as shown in \cref{fig:pre_compare}).

In short, the main contributions of this paper can be summarized as follows:

\begin{itemize}[leftmargin=2ex,topsep=0.25ex]
	\item We propose an iterative optimization algorithm named DMP, which can embed pre-trained diffusion models in the step of iterations.
	\item We provide the state evolution equation of DMP, which demonstrates its capacity to maintain data manifolds throughout the reconstruction process and explains the enhancement in reconstruction quality.
	\item We propose DMP-DUN by deep unfolding DMP, which can replace a large number of steps in the reverse diffusion process and Monte-Carlo SURE method steps with a lightweight network, thereby saving a significant amount of time. Moreover, DMP-DUN can obtain time steps and scaling parameters through end-to-end training, which are hyper-parameters in the original diffusion model, thus further improving the reconstruction process.
	\item Extensive experiments show that our proposed DMP-DUN can outperform current state-of-the-art methods in terms of reconstruction quality with lower computational costs, requiring only at least 2 steps of DMP.
\end{itemize}

\section{Related Work}
\subsection{Deep Unfolding Network}
By treating CS problem as a Lasso problem, many iterative algorithms such as ISTA\cite{Daubechies2003AnIT},  FISTA\cite{beck2009fast}, AMP\cite{Donoho2009MessagepassingAF}, and \etc\cite{Becker2009NESTAAF, afonso2010augmented}, have been developed in the past few decades. In recent years, with the development of deep learning, several network-based methods\cite{kulkarni2016reconnet, shi2017deep, xu2018lapran, shi2019scalable, ye2021csformer, fan2022global} have been proposed. Among them, a specific category known as Deep Unfolding Networks (DUNs)\cite{DBLP:conf/icml/GregorL10, DBLP:conf/nips/MetzlerMB17, yang2018admm, zhang2018ista, zhang2020amp, you2021coast, shen2022transcs, song2021memory, Song2023OptimizationInspiredCT, DBLP:journals/ijcv/ChenSXZ23, guoCPPNet2024, wangUFCNet2024, DBLP:journals/tmm/CuiFZZ24, DBLP:journals/tmm/KongCH24} unfold the traditional iterative algorithms and implement some of their functions through neural networks, thereby achieving higher reconstruction quality than other network-based methods.

Early DUNs\cite{DBLP:conf/icml/GregorL10, yang2018admm, zhang2018ista, zhang2020amp} learned prior knowledge, such as sparse domains, through end-to-end training, significantly enhancing convergence speed and reconstruction quality. With the development of DUNs, methods like \cite{song2021memory, Song2023OptimizationInspiredCT} addressed information loss during iterations. Additionally, several approaches\cite{you2021coast, shen2022transcs, DBLP:journals/ijcv/ChenSXZ23, guoCPPNet2024, wangUFCNet2024, DBLP:journals/tmm/CuiFZZ24, DBLP:journals/tmm/KongCH24} further improved reconstruction quality and robustness by refining network structures. For example, Shen \etal\cite{shen2022transcs} integrated Transformers, and Chen \etal\cite{DBLP:journals/ijcv/ChenSXZ23} designed a multi-scale gradient descent network.

The effectiveness of these methods in reconstruction largely depends on the quality of the prior knowledge they learn. Although some methods\cite{Yuan20PnPSCI, Zhou_2023_ICCV, Zheng:21} have been proposed to use pre-trained neural network as prior, there are no methods that utilize the powerful prior knowledge in pre-trained diffusion models.  
\subsection{Diffusion Model}
In recent years, the diffusion model\cite{DBLP:conf/nips/HoJA20, DBLP:conf/iclr/SongME21} has attracted attention from many researchers due to its excellent image generation quality and the modifiability of the reverse diffusion process. Several works\cite{DBLP:conf/nips/KawarEES22, wang2022zero, DBLP:conf/nips/ChungSRY22, he2024manifold, DBLP:conf/iclr/SongVMK23, DBLP:conf/iclr/ChungKMKY23, DBLP:conf/cvpr/FeiLPZYLZ023, DBLP:conf/nips/RoutRDCDS23, DBLP:journals/pami/SahariaHCSFN23} attempted to leverage pre-trained diffusion models to solve inverse problems, such as image CS, image denoising and super-resolution. These works, such as DDNM\cite{wang2022zero} and MPGD\cite{he2024manifold}, explored various strategies to adapt the diffusion process to the specific requirements of inverse problems, including modifying the diffusion steps, incorporating prior information, and optimizing the sampling techniques.

Meanwhile, some methods\cite{DBLP:journals/corr/abs-2403-17006, DBLP:journals/mia/ChungY22, DBLP:journals/tmi/CaoCWLCZLZ24, DBLP:conf/siu/MirzaC23} achieve higher quality reconstruction in natural image CS and MRI through redefining reverse diffusion process and retraining the diffusion model. Specifically, Chen \etal\cite{DBLP:journals/corr/abs-2403-17006} proposed the Invertible Diffusion Model (IDM), which implements end-to-end training of diffusion models, significantly improving the reconstruction quality and efficiency of diffusion models in image CS.

Compared to these methods, our proposed DMP-DUN can achieve higher quality reconstruction with fewer steps by unfolding our proposed DMP algorithm which embeds a diffusion model.

\section{Notation and Definitions} \label{sec:Notation}
This paper uses mathematical boldface to represent tensors, with uppercase boldface for matrices. The normal distribution is denoted by $\mathcal{N}$, and the identity matrix by $\I$. The sets of real numbers and natural numbers are denoted by $\R$ and $\mathbb{N}$, respectively.
	
Traditional iterative optimization algorithms start at $t=0$ and converge to the theoretical optimal solution at $t=+\infty$. In contrast, the reconstruction process of diffusion models begins at $t=+\infty$ and results in the reconstructed image at $t=0$. Therefore, to ensure that the iterative processes of the two methods match, this paper standardizes $t=+\infty$ as the starting point of iteration and $t=0$ as the end point.
	
We employ the block-based compressive sensing method\cite{trevisi2019compressive, zhang2014group}, which divides the image into non-overlapping blocks of size $B \times B$ and flattens each block, where $B^2 = N$. For simplicity, we implicitly assume that the image has been divided and flattened into a one-dimensional tensor when performing left-multiplication matrix throughout this paper.

\begin{figure*}[t]
	\centering
	\includegraphics[width=\linewidth]{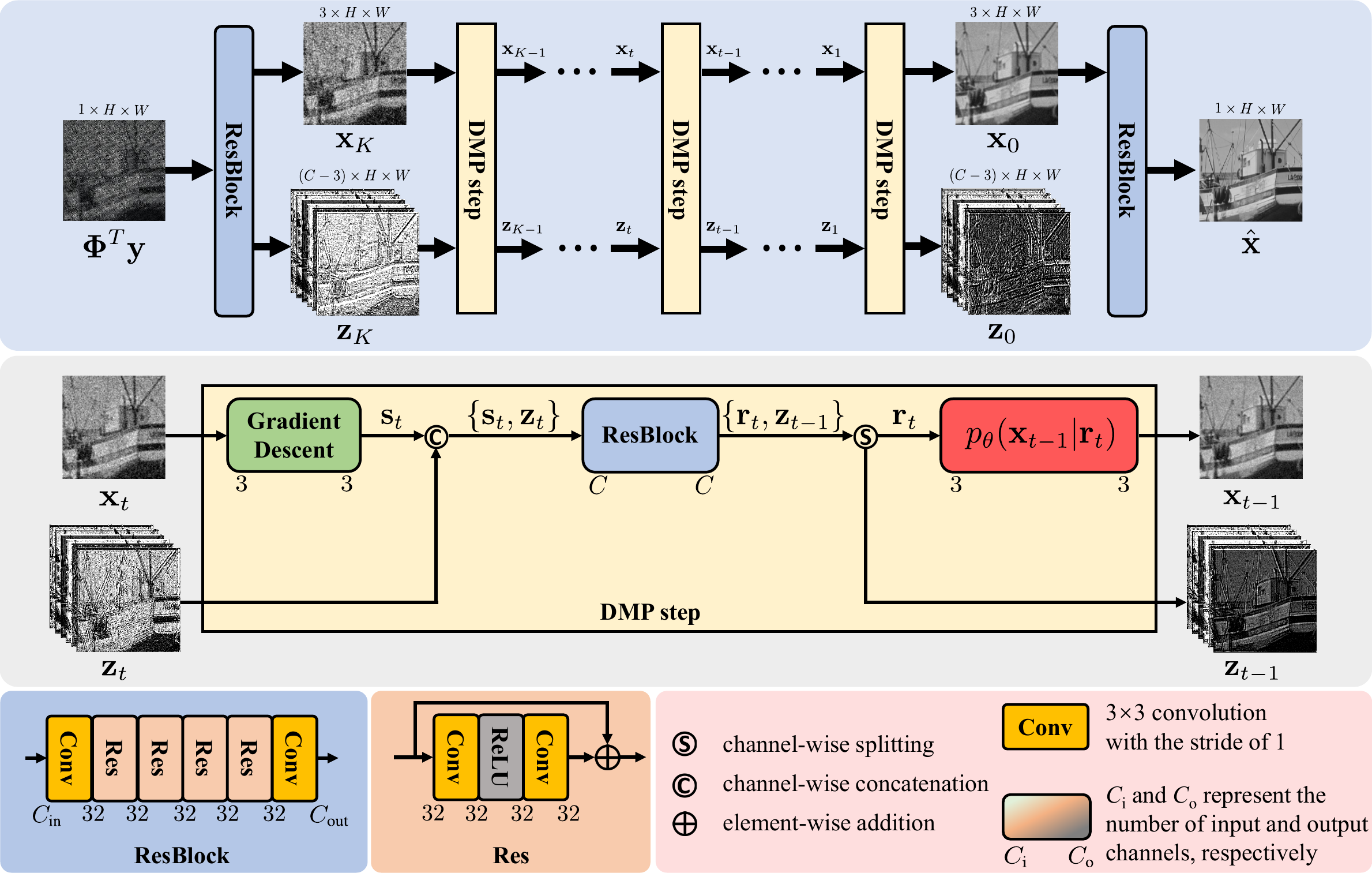}
	\caption{The reconstruction process of our proposed DMP-DUN, where DMP step represents each iteration of Diffusion Message Passing (DMP) with pre-trained Diffusion Model (\ie $p_\theta(\x_{t-1}|\rr_t)$).}
	\label{fig:network_summary_detail}
\end{figure*}
\section{Method} \label{sec:Method}

\subsection{Preliminaries} \label{subsec:Preliminaries}
The original image $\mathbf{x} \in \mathbb{R}^N$ is obtained through the sensing matrix $\mathbf{\Phi} \in \mathbb{R}^{M \times N}$, resulting in CS measurements $\mathbf{y} \in \mathbb{R}^M$, where $M \ll N$. The problem of reconstructing $\mathbf{x}$ from the known $\mathbf{y}$ and $\mathbf{\Phi}$ can be formulated as the following optimization problem:
\begin{equation}
	\hat{\x } = \mathop{\arg\min}_{\x} \| \y - \Phix \|_2^2 \qquad \text{s.t. } \x \in \mathcal{X}
\end{equation}
where $\| \cdot \|_2$ means $\ell_2$ norm, $\mathcal{X} \subset \mathbb{R}^N$ represents a set constrained by certain prior knowledge. For example, $\mathcal{X}$ can be defined as the collection of all sparse representations in a specific transform domain (such as the DCT domain) for natural images. In addressing this problem, AMP makes the following assumptions:
\begin{assumption} \label{assum:amp_infty}
	The Gaussian sensing matrix $\mathbf{\Phi} \sim \mathcal{N}(0, \I/M)$. The sensing rate $\delta = M/N \in \mathbb{R} (0, 1)$, where $M, N \to +\infty$. The denoising function $\eta_t(\cdot)$ is Lipschitz continuous, with $t\in\mathbb{N}$. 
\end{assumption}

Based on the above assumption, AMP gives the following iterative formula to reconstruct the image with the iteration starts from $t = +\infty$ and ends when $t = 0$, and the final answer $\x_0$ will be converged to ground truth:
\begin{equation} \label{eq:amp}
	\begin{aligned}
		\bfu_t &= \y - \Phix_t + \bfu_{t+1} \text{div}\eta_t(\hh_{t+1})  / M \\
		\hh_t &= \PhiTu_t + \x_t \\
		\x_{t-1} &= \eta_t(\hh_t)
	\end{aligned}
\end{equation}
where $\text{div}(\cdot)$ represents the divergence and the term $[\bfu_{t+1} \text{div}\eta_t(\hh_{t+1})/M]$ in AMP is referred to as the Onsager term. If the denoiser $\eta_t (\cdot)$ is a black-box function such as a deep neural network, then its divergence can be approximated through the method of Monte Carlo approximation\cite{DBLP:journals/tip/RamaniBU08, DBLP:journals/tit/MetzlerMB16}, that is, by using $\mathbf{g}[\eta_t(\hh_{t+1}+\varepsilon \mathbf{g}) - \eta_t(\hh_{t+1})]/\varepsilon$ to approximate divergence of $\eta_t (\cdot)$, where $\mathbf{g} \sim \mathcal{N} (0, \I)$ and $\varepsilon \to 0$. Onsager term can decouple the errors across iterations, resulting in the iterative formula that approximately follows the following state evolution:
\begin{equation}
	\begin{aligned}
		\hh_t &\sim \mathcal{N} (\x, \sigma^2_t \I / \delta) \\
		\sigma^2_{t-1} &= \mathbb{E}\left\{ [\eta_t(\hh_t) - \x]^2 \right\}
	\end{aligned}
\end{equation}

\subsection{Diffusion Message Passing Algorithm}
In this sub-section, we will elaborate how the Diffusion Message Passing (DMP) algorithm embeds a pre-trained diffusion model into its iterations to harness prior knowledge.

In the diffusion model\cite{DBLP:conf/nips/HoJA20, DBLP:conf/iclr/SongME21}, for the original image $\x$, let $\alpha_t\in \mathbb{R}[0, 1]$ decrease with increasing $t\in\mathbb{N}$, where the scaling parameter $\alpha_0=1$. The forward diffusion process start at $\x_0 = \x$, and at the $t$-th step is given by $q(\x_t | \x_{t-1}) = \mathcal{N} (\sqrt{\alpha_t} \x_{t-1}, (1-\alpha_t)\I)$. The parameter $\alpha_t=0$ when $t = +\infty$, so at this time $\x_t\sim \mathcal{N}(\mathbf{0}, \I)$. Under the assumption that $\x_0$ is known, the denosing diffusion model can obtain the reverse diffusion process as follows:
\begin{equation}
	q(\x_{t-1}|\x_t,\x_0) = \mathcal{N}(\tilde{\mu_t}(\x_t,\x_0),\tilde{\beta_t}\I)
\end{equation}
where $\tilde{\mu}(\x_t,\x_0) = \frac{1-\bar{\alpha}_{t-1}}{1-\bar{\alpha}_t}\sqrt{\alpha_t}\x_t+\frac{1-\alpha_t}{1-\bar{\alpha}_t}\sqrt{\bar{\alpha}_{t-1}}\x_0$, $\tilde{\beta_t}=\frac{1-\bar{\alpha}_{t-1}}{1-\bar{\alpha}_t}(1 - \alpha_t)$, and $\bar{\alpha}_t = \prod_{i=1}^{t}\alpha_i$. For $\x_0$ is unknown  in practical scenarios, diffusion model use $p_\theta(\x_{t-1}|\x_t)$ to approximate $q(\x_{t-1}|\x_t,\x_0)$, where $\theta$ represents the learnable parameters in the network.

\begin{assumption} \label{assum:ideal_diffusion}
	The neural network $p_\theta(\x_{t-1}|\x_t)$ is a perfect denoiser, meaning $p_\theta(\x_{t-1}|\x_t) \equiv q(\x_{t-1}|\x_t,\x_0)$. Consequently, $\x_t \sim\mathcal{N} (\sqrt{\bar{\alpha}_t} \x, (1-\bar{\alpha}_t)\I)$ for all $t\in\mathbb{N}$.
\end{assumption}
\begin{assumption} \label{assum:ideal_filter}
	A perfect Gaussian filter exists and is represented by $\mathcal{D}_t(\cdot)$. The distribution $q(\mathcal{D}_t( \bfd_t ) | \bfd_t) = \mathcal{N}(\sqrt{\bar{\alpha}_t}\x, (1-\sqrt{\bar{\alpha}_t})\I)$, where $\bfd_t\sim \mathcal{N}(\sqrt{\bar{\alpha}_t}\x, (1-\sqrt{\bar{\alpha}_t})\I/\delta)$.
\end{assumption}

Based on the aforementioned description and considering the diffusion model as a denoiser, we can derive the following theorem for our proposed DMP algorithm:
\begin{theorem} \label{theorem:amp_for_ddm}
	Suppose Assumption \ref{assum:amp_infty}, \ref{assum:ideal_diffusion}, and \ref{assum:ideal_filter} hold, we deduce the iterative representation of Diffusion Message Passing (DMP) algorithm as follows:
	\begin{equation} \label{eq:ddm_amp}
		\begin{aligned}
			\bfs_t &= \x_t - \sqrt{\barAlpha_t} \bfPhi^\mathrm{T}(\Phix_t - \y) \\
			\rr_t &= \mathcal{D}_t[ \bfs_t + \sqrt{\barAlpha_t} o_t(\bfu_{t+1}, \hh_{t+1})]\\
			\x_{t-1} &= p_\theta(\x_{t-1}|\rr_t)
		\end{aligned}
	\end{equation}
	where $o_t(\bfu_{t+1}, \hh_{t+1}) := \bfPhi^\mathrm{T} \bfu_{t+1} \mathrm{div}\eta_t(\hh_{t+1})$. Moreover, it is possible to write out its state evolution:
	\begin{equation} \label{eq:ddm_amp_se}
		\begin{aligned}
			\rr_t &\sim \mathcal{N} (\sqrt{\barAlpha_t}\x, \sigma^2_t \I) \\
			\sigma^2_{t-1} &= \mathbb{E}\left\{ [p_\theta(\x_{t-1}|\rr_t) - \sqrt{\barAlpha_{t-1}}\x]^2 \right\}
		\end{aligned}
	\end{equation}
\end{theorem}

Due to space constraints, we relegate the detailed proof to Supplementary Materials for Theorem \ref{theorem:amp_for_ddm}.

\subsection{DMP-DUN} \label{subsec:ddamp}
Although Theorem \ref{theorem:amp_for_ddm} can be derived under the Assumptions \ref{assum:amp_infty}, \ref{assum:ideal_diffusion}, and \ref{assum:ideal_filter}, it is hard to fully meet these assumptions in practical problems. Furthermore, the term $o_t(\bfu_{t+1}, \hh_{t+1})$ in \cref{eq:ddm_amp} necessitates the computation of divergence of the denoiser. When the denoiser is a black box, it becomes imperative to employ the Monte-Carlo SURE\cite{DBLP:journals/tip/RamaniBU08} method for resolution, which thus doubles the computational expense relative to the denoiser itself. Considering the success of DUN in the CS field, we propose a novel method named DMP-DUN in this paper, which deeply unfolds DMP algorithm and introduces several convolutional residual blocks (ResBlocks) to implement functions such as $o_t$ and $\mathcal{D}_t$ within \cref{eq:ddm_amp}. Moreover, DMP-DUN enables the learning of what were traditionally hyper-parameters like $\barAlpha_t$ and $t$, which previously required manual tuning in diffusion models. 

The specific structure of DMP-DUN is illustrated in \cref{fig:network_summary_detail}, where each DMP step denotes an iterative step of DMP algorithm (\ie \cref{eq:ddm_amp}). Specifically, the Gradient Descent, ResBlock, and $p_\theta(\x_{t-1}|\rr_t)$ in DMP represent the steps for obtaining $\bfs_t$, $\rr$, and $\x_{t-1}$ in \cref{eq:ddm_amp}, respectively. Additionally, We place the ResBlock at the head and the tail of DMP-DUN, respectively. The ResBlock at head of DMP-DUN can save the time required for the reverse diffusion process from $T$ to $K$ by processing the input image $\PhiTy$ into an image $\x_K$ that conforms to the distribution of the diffusion model at time $K$, where the $K$ is less than the start time $T$ of the original diffusion model. The feature map $\zz_t$ can be represented as $\zz_t (\bfu_{t+1}, \hh_{t+1})$ according to the definitions in \cref{eq:amp} and (\ref{eq:ddm_amp}) and can be applied in the subsequent DMP step, where the initial feature map $\zz_K$ is also generated by the ResBlock at head of DMP-DUN. The ResBlock at the end of DMP-DUN functions to transform the final RGB image output of the DMP into a single-channel image, thereby constructing the bridge between the RGB image processed by a pre-trained diffusion model and the Y-channel image that needs to be reconstructed.

\begin{table*}[th]
	\centering
	\begin{adjustbox}{width=\linewidth,center}
		\begin{tabular}{@{}lrccccccl@{}}
			\toprule
			\multicolumn{2}{c}{\multirow{2}{*}{Method}} & \multicolumn{5}{c}{Set11\cite{kulkarni2016reconnet}} & \multirow{2}{*}{Avg.} & \multicolumn{1}{c}{FLOPs} \\ \cmidrule(lr){3-7}
			\multicolumn{2}{c}{} & 1\% & 4\% & 10\% & 25\% & 50\% &  & \multicolumn{1}{c}{(G)} \\ \midrule
			ISTA-Net$^+$\cite{zhang2018ista} & (CVPR 2018) & 17.45/0.4131 & 21.56/0.6240 & 26.49/0.8036 & 32.44/0.9237 & 38.08/0.9680 & 27.20/0.7465 & 56.2 \\
			AMP-Net\cite{zhang2020amp} & (TIP 2021) & 20.20/0.5581 & 25.26/0.7722 & 29.40/0.8779 & 34.63/0.9481 & 40.34/0.9807 & 29.97/0.8274 & 44.9 \\
			TransCS\cite{shen2022transcs} & (TIP 2022) & 20.15/0.5066 & 25.41/0.7883 & 29.54/0.8877 & 35.06/0.9548 & 40.21/0.9824 & 29.02/0.8174 & 26.6 \\
			CSformer\cite{ye2021csformer} & (TIP 2023) & 21.58/0.6075 & 26.28/0.8062 & 29.79/0.8883 & 34.81/0.9527 & 40.73/0.9824 & 31.21/0.8584 & 72.9 \\
			DPC-DUN\cite{DBLP:journals/tip/SongCZ23} & (TIP 2023) & 18.03/0.4601 & 24.38/0.7498 & 29.42/0.8801 & 34.75/0.9483 & 39.84/0.9778 & 29.28/0.8032 & 138.0 \\
			OCTUF\cite{Song2023OptimizationInspiredCT} & (CVPR 2023) & 21.75/0.5934 & 26.45/0.8126 & 30.70/0.9030 & 36.10/0.9604 & 41.34/0.9838 & 31.27/0.8506 & 189.3 \\
			PRL-RND$^+$\cite{DBLP:journals/ijcv/ChenSXZ23} & (IJCV 2023) & \underline{22.27}/\underline{0.6240} & 26.88/0.8130 & \underline{31.66}/\underline{0.9189} & \underline{36.69}/0.9608 & \underline{41.84}/\underline{\textcolor{blue}{0.9850}} & \underline{31.87}/0.8603 & 562.2 \\
			CAT-Net\cite{DBLP:journals/tmm/KongCH24} & (TMM 2024) & 21.29/0.5782 & 26.38/0.8060 & 30.69/0.9022 & 35.85/0.9588 & 41.28/0.9834 & 31.10/0.8457 & 94.4 \\
			UFC-Net\cite{wangUFCNet2024} & (CVPR 2024) & 21.24/0.5607 & 25.92/0.7943 & 30.15/0.8960 & 35.42/0.9567 & - & - & 109.0 \\
			CPP-Net\cite{guoCPPNet2024} & (CVPR 2024) & 22.19/0.6135 & \underline{27.23}/\underline{0.8337} & 31.27/0.9135 & 36.35/\underline{0.9631} & 41.39/0.9827 & 31.69/\underline{0.8613} & 153.5 \\
			DDNM$_\text{(10-step)}$\cite{wang2022zero} & (ICLR 2023) & 11.89/0.1748 & 12.29/0.1338 & 13.39/0.1404 & 16.98/0.2959 & 25.63/0.8074 & 16.04/0.3105 & 670.4 \\
			DDNM$^+_\text{(1000-step)}$\cite{wang2022zero} & (ICLR 2023) & 17.95/0.4497 & 22.54/0.6827 & 25.78/0.8154 & 27.80/0.8933 & 29.01/0.9411 & 24.62/0.7564 & 67039.1 \\ \midrule
			DMP-DUN$_\text{(10-step)}$ & (ours) & \textcolor{blue}{23.32}/\textcolor{blue}{0.6305} & 28.20/0.8340 & 32.51/0.9161 & \textcolor{blue}{37.92}/\textcolor{blue}{0.9668} & \textcolor{red}{42.99}/\textcolor{red}{0.9857} & \textcolor{blue}{32.99}/0.8666 & 742.6 \\
			DMP-DUN$^+_\text{(2-step)}$ & (ours) & 23.18/0.6286 & \textcolor{blue}{28.25}/\textcolor{blue}{0.8360} & \textcolor{blue}{32.63}/\textcolor{blue}{0.9206} & 37.58/0.9648 & 42.06/0.9835 & 32.74/\textcolor{blue}{0.8667} & 157.0 \\
			DMP-DUN$^+_\text{(4-step)}$ & (ours) & \textcolor{red}{23.32}/\textcolor{red}{0.6313} & \textcolor{red}{28.67}/\textcolor{red}{0.8448} & \textcolor{red}{33.22}/\textcolor{red}{0.9277} & \textcolor{red}{38.29}/\textcolor{red}{0.9681} & \textcolor{blue}{42.82}/0.9848 & \textcolor{red}{33.26}/\textcolor{red}{0.8713} & 303.4 \\ \bottomrule
		\end{tabular}
	\end{adjustbox}
	
	\begin{adjustbox}{width=\linewidth,center}
		\begin{tabular}{@{}lrccccccl@{}}
			\toprule
			\multicolumn{2}{c}{\multirow{2}{*}{Method}} & \multicolumn{5}{c}{Urban100\cite{Dong2018DenoisingPD}} & \multirow{2}{*}{Avg.} & \multicolumn{1}{c}{FLOPs} \\ \cmidrule(lr){3-7} 
			\multicolumn{2}{c}{} & 1\% & 4\% & 10\% & 25\% & 50\% &  & \multicolumn{1}{c}{(G)} \\ \midrule
			ISTA-Net$^+$\cite{zhang2018ista} & (CVPR 2018) & 16.66/0.3733 & 19.65/0.5368 & 23.48/0.7200 & 28.89/0.8830 & 34.43/0.9571 & 24.62/0.6940 & 56.2 \\
			AMP-Net\cite{zhang2020amp} & (TIP 2021) & 19.55/0.5016 & 22.73/0.6819 & 25.92/0.8144 & 30.79/0.9188 & 36.33/0.9712 & 27.06/0.7776 & 44.9 \\
			TransCS\cite{shen2022transcs} & (TIP 2022) & 18.96/0.4395 & 23.25/0.7114 & 26.74/0.8416 & 31.75/0.9329 & 37.20/0.9761 & 27.58/0.7803 & 26.6 \\
			CSformer\cite{ye2021csformer} & (TIP 2023) & \underline{21.57}/\underline{0.5672} & \underline{24.94}/0.7396 & 27.92/0.8458 & 32.43/0.9332 & 37.88/0.9766 & 28.95/0.8125 & 72.9 \\
			DPC-DUN\cite{DBLP:journals/tip/SongCZ23} & (TIP 2023) & 17.28/0.4214 & 22.35/0.6767 & 26.94/0.8358 & 32.33/0.9320 & 37.52/0.9737 & 27.28/0.7679 & 138.0 \\
			OCTUF\cite{Song2023OptimizationInspiredCT} & (CVPR 2023) & 19.88/0.5167 & 23.68/0.7329 & 27.79/0.8621 & 32.99/0.9445 & 38.29/0.9797 & 28.53/0.8072 & 189.3 \\
			PRL-RND$^+$\cite{DBLP:journals/ijcv/ChenSXZ23} & (IJCV 2023) & 20.63/0.5632 & 24.32/0.7439 & \underline{28.75}/\underline{0.8743} & \underline{34.00}/0.9483 & \underline{38.64}/\underline{0.9798} & \underline{29.27}/0.8219 & 562.2 \\
			CAT-Net\cite{DBLP:journals/tmm/KongCH24} & (TMM 2024) & 19.63/0.5209 & 23.64/0.7188 & 27.59/0.8485 & 32.48/0.9340 & 37.77/0.9751 & 28.22/0.7995 & 94.4 \\
			UFC-Net\cite{wangUFCNet2024} & (CVPR 2024) & 19.69/0.5041 & 23.37/0.7195 & 27.55/0.8583 & 32.82/0.9423 & - & - & 109.0 \\
			CPP-Net\cite{guoCPPNet2024} & (CVPR 2024) & 20.55/0.5554 & 24.66/\underline{0.7691} & 28.49/0.8801 & 33.38/\underline{0.9485} & 38.33/0.9781 & 29.08/\underline{0.8262} & 153.5 \\
			DDNM$_\text{(10-step)}$\cite{wang2022zero} & (ICLR 2023) & 10.53/0.1374 & 11.00/0.1210 & 12.26/0.1441 & 16.28/0.3293 & 24.81/0.8147 & 14.98/0.3093 & 670.4 \\
			DDNM$^+_\text{(1000-step)}$\cite{wang2022zero} & (ICLR 2023) & 17.48/0.4114 & 21.86/0.6435 & 24.62/0.7880 & 26.22/0.8621 & 27.35/0.9118 & 23.51/0.7234  & 67039.1 \\ \midrule
			DMP-DUN$_\text{(10-step)}$ & (ours) & 21.48/0.5671 & 25.80/0.7727 & 30.04/0.8857 & \textcolor{blue}{35.25}/\textcolor{blue}{0.9538} & \textcolor{blue}{40.44}/\textcolor{blue}{0.9827} & \textcolor{blue}{30.60}/0.8324 & 742.6 \\
			DMP-DUN$^+_\text{(2-step)}$ & (ours) & \textcolor{blue}{21.66}/\textcolor{blue}{0.5750} & \textcolor{blue}{26.25}/\textcolor{blue}{0.7858} & \textcolor{blue}{30.41}/\textcolor{blue}{0.8922} & 35.07/0.9520 & 39.46/0.9795 & 30.57/\textcolor{blue}{0.8369} & 157.0 \\
			DMP-DUN$^+_\text{(4-step)}$ & (ours) & \textcolor{red}{21.80}/\textcolor{red}{0.5832} & \textcolor{red}{26.98}/\textcolor{red}{0.8035} & \textcolor{red}{31.39}/\textcolor{red}{0.9053} & \textcolor{red}{36.14}/\textcolor{red}{0.9584} & \textcolor{red}{40.80}/\textcolor{red}{0.9827} & \textcolor{red}{31.42}/\textcolor{red}{0.8466} & 303.4 \\ \bottomrule
		\end{tabular}
	\end{adjustbox}
	\caption{Average PSNR/SSIM performance comparisons of various CS methods on Set11\cite{kulkarni2016reconnet} and Urban100\cite{Dong2018DenoisingPD}. The best and second best results are highlighted in \textcolor{red}{red} and \textcolor{blue}{blue} colors, respectively. The best results among the compared methods are highlighted with \underline{underlines}. }
	\label{table:main_ex_urban100}
\end{table*}

In the Gradient Descent module, we use a learnable parameter $\lambda_t\in\mathbb{R}^+$ to replace $\sqrt{\bar{\alpha}}_t$ in \cref{eq:ddm_amp} to increase fault tolerance, and the formula is as follows:
\begin{equation}
	\bfs_t = \x_t - \lambda_t \bfPhi^\mathrm{T}(\Phix_t - \y)
\end{equation}

To obtain the $\rr_t$ in \cref{eq:ddm_amp}, we proposed to directly use a lightweight neural network to fit the $\mathcal{D}_t$ and learn how to get the onsager term directly by using the ResBlock: 
\begin{equation}
		\{ \rr_t, \zz_{t-1} \} =\mathcal{H}_\text{Conv} \left( \mathcal{H}_\text{4-Res} \left( \mathcal{H}_\text{Conv} \left( \{ \bfs_t, \zz_t \} \right) \right) \right)
\end{equation}
where $\mathcal{H}_\text{Conv}$ is a $3\times 3$ convolution with stride is 1, and $\mathcal{H}_\text{4-Res}$ denotes 4 cascaded Res structures. Each Res structure use $\x + \mathcal{H}_\text{Conv} (\text{ReLU}(\mathcal{H}_\text{Conv}(\x))) $ to implement. The resultant output $\mathbf{z}_{t-1}$ is utilized in the subsequent iteration of the DMP step.

Subsequently, $\mathbf{r}_t$ serves as the input for the ensuing module $p_\theta (\mathbf{x}_{t-1}|\mathbf{r}_t)$, which is implemented using a pre-trained diffusion model. Such integration can leverage powerful prior knowledge of the pre-trained diffusion model, and can compatible with various diffusion models.

\subsection{DMP-DUN$^+$}
Instead of freezing the parameters of diffusion model when training DMP-DUN, the proposed DMP-DUN$^+$ allows them to be unfrozen during training. This approach will significantly increase the training cost but can further improve the reconstruction quality. Additionally, this operation only affects the training cost and does not impact the cost of using it for reconstruction.

\subsection{End-to-End Learning} \label{subsec:e2e}
Unlike diffusion models which trained to minimize the local reverse diffusion process between $q(\x_{t-1}|\x_t, \x_0)$ and $p_\theta(\x_{t-1}|\x_t)$ as the optimization target, our proposed DMP-DUN and DMP-DUN$^+$ uses the training strategy of DUNs that employs end-to-end training to directly optimize the entire reconstruction process. Given a set of full-sampled images $\{\mathbf{x}_i\}_{i=1}^{N_a}$ and its CS measurements set $\{\y_i|\y_i = \Phix_i\}_{i=1}^{N_a}$, we use the mean squared error (MSE) as the loss function. We update all parameters in DMP-DUN except for the parameters in pre-trained diffusion model. For DMP-DUN$^+$, we update all parameters including those of diffusion model. The training objective of DMP-DUN and DMP-DUN$^+$ can be represented as shown below:
\begin{equation}
	\min\limits_\mathbf{\Theta} \frac{1}{N_p N_a}\sum^{N_a}_{i=1}\| \mathcal{F}_\mathbf{\Theta}(\PhiTy_i) - \x_i \|_2^2 \label{eq:loss} 
\end{equation}
where $\mathcal{F}_\mathbf{\Theta}$ denotes the entire reconstruction process of DMP-DUN or DMP-DUN$^+$, $N_a$ is the size of training dataset, $N_p$ is the number of pixels in each image $\mathbf{x}_i$. $\mathbf{\Theta}$ denotes the set of all trainable parameters of DMP-DUN or DMP-DUN$^+$, which includes the trainable sensing matrix $\bfPhi$.

\begin{figure*}[t]
	\centering
	\includegraphics[width=\linewidth]{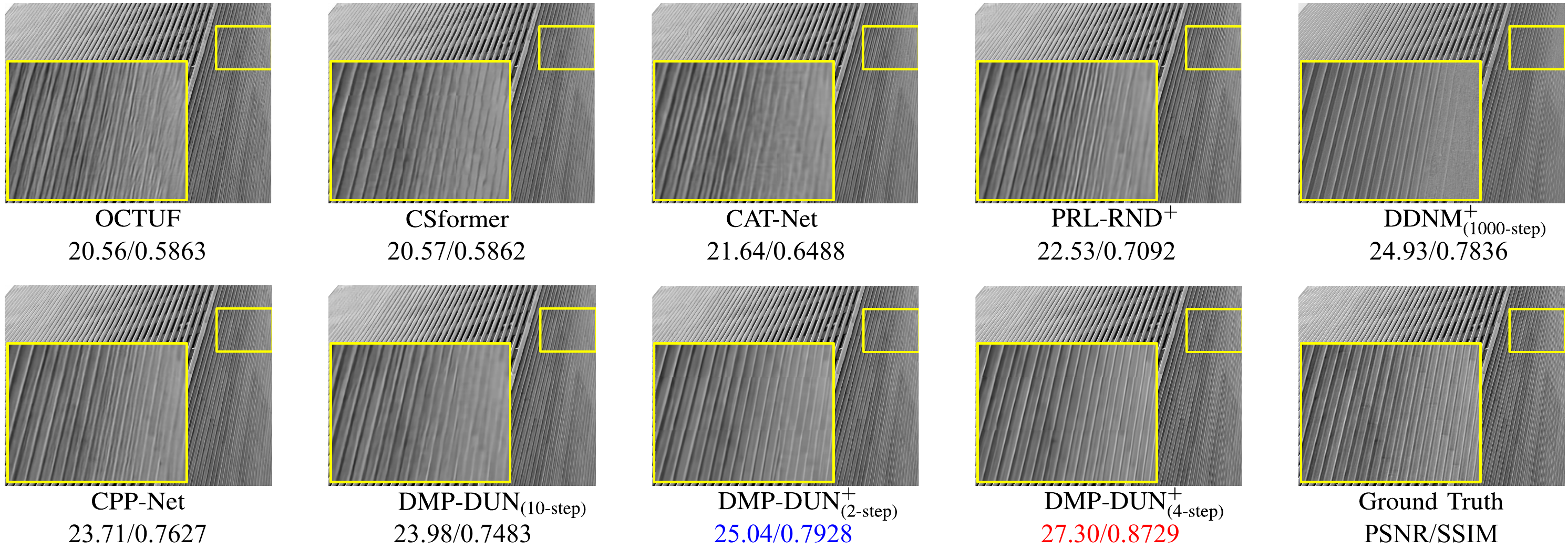}
	\caption{The comparison of the subjective visual effects between our proposed DMP-DUN, DMP-DUN$^+$, and other methods at CS ratio = 4\%. The best and second best results are highlighted in \textcolor{red}{red} and \textcolor{blue}{blue} colors, respectively.
	}
	\label{fig:reconstruction_compare}
\end{figure*}

\begin{table}[t]
	\centering
	\begin{adjustbox}{width=\linewidth,center}
		\begin{tabular}{@{}lccccccc@{}}
			\toprule
			\multicolumn{1}{c}{} & \multicolumn{3}{c}{Urban100\cite{Dong2018DenoisingPD}} & \multicolumn{3}{c}{DIV2K\cite{Agustsson_2017_CVPR_Workshops}} &  \\ \cmidrule(lr){2-4}\cmidrule(lr){5-7}
			\multicolumn{1}{c}{\multirow{-2}{*}{Method}} & 10\% & 30\% & 50\% & 10\% & 30\% & 50\% & \multirow{-2}{*}{Avg.} \\ \midrule
			DDNM$^+$ (ICLR'23) & 24.61 & 26.74 & 27.71 & 24.40 & 26.61 & 27.73 & 26.30 \\
			IDM~~~~~(TPAMI'25) & \textcolor{blue}{31.41} & \textcolor{blue}{36.76} & \textcolor{blue}{40.33} & \textcolor{blue}{31.07} & \textcolor{blue}{36.98} & \textcolor{blue}{41.15} & \textcolor{blue}{36.28} \\ 
			DMP-DUN$^+$~ (ours) & \textcolor{red}{31.90} & \textcolor{red}{37.45} & \textcolor{red}{41.58} & \textcolor{red}{32.56} & \textcolor{red}{38.32} & \textcolor{red}{42.66} & \textcolor{red}{37.41} \\ \bottomrule
		\end{tabular}
	\end{adjustbox}
	\caption{Average PSNR performance comparisons of various methods based on diffusion model with $256\times 256$ center-cropped.}
	\label{table:main_ex_idm}
\end{table}

\begin{table*}[t]
	\begin{adjustbox}{width=0.75\linewidth,center}
		\begin{tabular}{@{}lcccccc@{}}
			\toprule
			\multicolumn{1}{c}{\multirow{2}{*}{Method}} & \multirow{2}{*}{Pre-trained} & \multicolumn{2}{c}{Set11} & \multicolumn{2}{c}{Urban100} & \multirow{2}{*}{Avg.} \\ \cmidrule(lr){3-4}\cmidrule(lr){5-6}
			\multicolumn{1}{c}{} &  & 4\% & 25\% & 4\% & 25\% &  \\ \midrule
			CSformer\cite{ye2021csformer} & \XSolidBrush & 26.28/0.8062 & 34.81/0.9527 & 24.94/0.7396 & 32.43/0.9332 & 29.62/0.8579 \\
			PRL-RND$^+$\cite{DBLP:journals/ijcv/ChenSXZ23} & \XSolidBrush & 26.88/0.8130 & 36.69/0.9608 & 24.32/0.7439 & 34.00/0.9483 & 30.47/0.8665 \\
			CPP-Net\cite{guoCPPNet2024} & \XSolidBrush & 27.23/0.8337 & 36.35/0.9631 & 24.66/0.7691 & 33.38/0.9485 & 30.41/0.8786 \\ \midrule
			DMP-DUN$^+_\text{(4-step)}$ & \Checkmark & 28.67/0.8448 & 38.29/0.9681 & 26.98/0.8035 & 36.14/0.9584 & 32.52/0.8937 \\
			DMP-DUN$^+_\text{(4-step)}$ & \XSolidBrush & 28.43/0.8407 & 38.07/0.9672 & 26.28/0.7878 & 35.63/0.9557 & 32.10/0.8879 \\ \bottomrule
		\end{tabular}
	\end{adjustbox}
	\caption{Comparison of DMP-DUN$^+$ with and without loading the pre-trained parameter at the begin of training.}
	\label{table:pretrained_ex}
\end{table*}

\begin{table*}[t]
	\centering
	
	\begin{adjustbox}{width=0.85\linewidth,center}
		\begin{tabular}{@{}cccccccc@{}}
			\toprule
			\multicolumn{1}{c}{Case} & Gradient Descent & ResBlock & $p_\theta (\mathbf{x}_{t-1}|\mathbf{r}_t)$ & Set11 & Set14 & Urban100 & Avg. \\ \midrule
			(a) & \XSolidBrush & \Checkmark & \Checkmark & 29.89/0.8713 & 28.77/0.8015 & 27.27/0.8225 & 28.64/0.8318 \\
			(b) & \Checkmark & \XSolidBrush & \Checkmark & 32.14/0.9097 & 31.04/0.8476 & 29.73/0.8702 & 30.97/0.8758 \\
			(c) & \Checkmark & \Checkmark & \XSolidBrush & 31.89/0.9009 & 30.82/0.8423 & 29.14/0.8627 & 30.62/0.8686 \\
			\rowcolor[HTML]{EFEFEF} 
			DMP-DUN$^+_\text{(4-step)}$ & \Checkmark & \Checkmark & \Checkmark & 33.22/0.9277 & 31.71/0.8664 & 31.39/0.9053 & 32.11/0.8998 \\ \bottomrule
		\end{tabular}
	\end{adjustbox}
	\caption{Ablation study of DMP-DUN at CS ratio = 10\%.}
	\label{table:abl_ex}
\end{table*}

\begin{figure}[t]
	\centering
	\includegraphics[width=\linewidth]{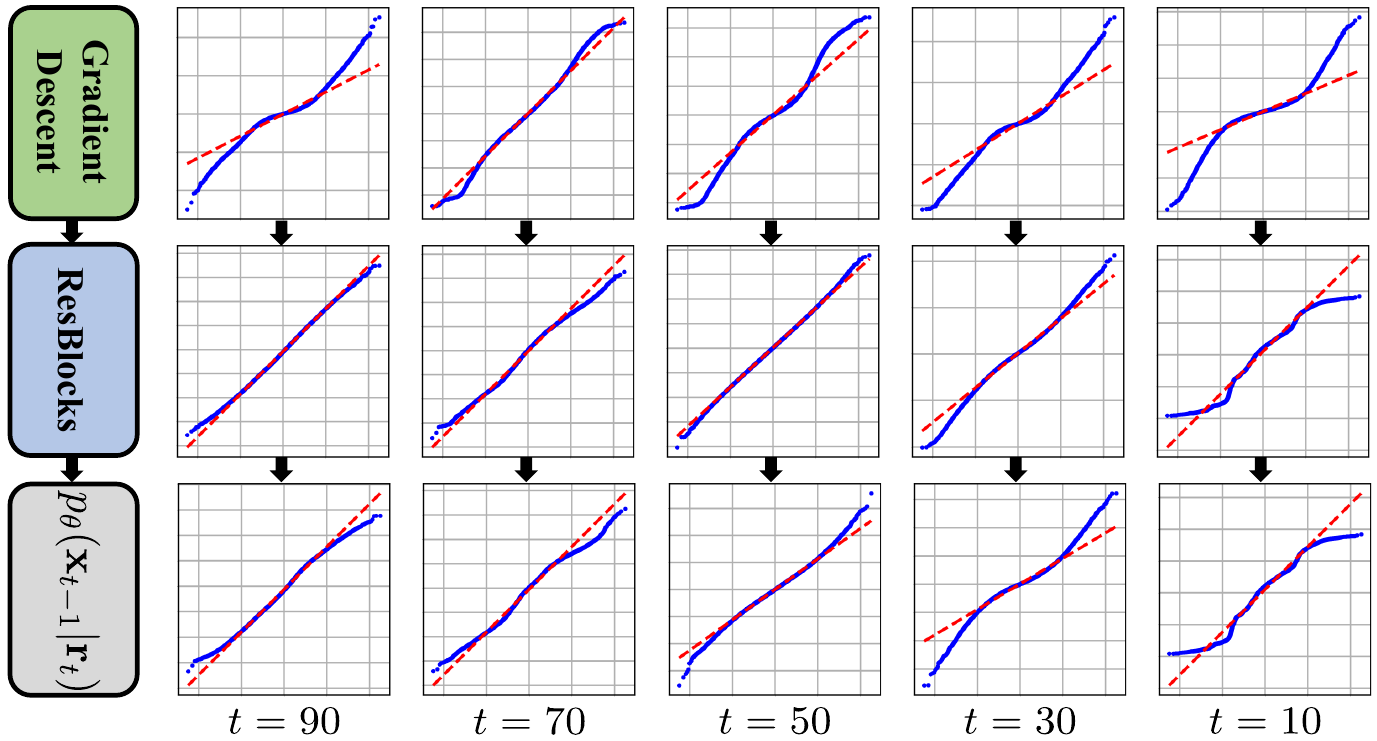}
	\caption{Q-Q plots for \textbf{the noise in output images of each module in DMP step} at different $t$. The horizontal axis represents the quantiles of normal distribution and the vertical axis represents the quantiles of noise. The closer the distribution of the \textcolor{blue}{blue points} is to the \textcolor{red}{red line}, the more the noise conforms to a normal distribution.}
	\label{fig:qq_plot}
\end{figure}

\begin{table}[t]
	\centering
	\begin{adjustbox}{width=1.0\columnwidth,center}
		\begin{tabular}{@{}lcccc@{}}
			\toprule
			Steps & 2 & 4 & 5 & 10 \\ \midrule
			Set11 & 31.39/0.9030 & 32.02/0.9091 & 32.12/0.9106 & 32.51/0.9161 \\
			Urban100 & 28.47/0.8600 & 29.28/0.8728 & 29.57/0.8776 & 30.04/0.8857 \\ \bottomrule
		\end{tabular}
	\end{adjustbox}
	\caption{Average PSNR/SSIM performance comparisons of various iterative steps of DMP-DUN under CS ratio = 10\%.}
	\label{table:iterative_steps}
\end{table}

\section{Experiment}

\subsection{Experimental Settings} \label{subsec:ex_setting}
We set the block size $B=64$, and the number of measurements $M=\lfloor \delta N \rfloor$, where $\delta \in \{ 1\%, 4\%, 10\%, 25\%, 50\% \}$ is the CS ratio. The channel numbers $C$ is set as $32$. The DMP-DUN and DMP-DUN$^+$ employ the technique from DDIM\cite{DBLP:conf/iclr/SongME21} to accelerate the reverse diffusion process with initial time $K=100$. We use the pre-trained diffusion model on $256 \times 256$ image in ImageNet dataset by \cite{DBLP:conf/nips/DhariwalN21}.  According to different acceleration step size $\Delta t$, we propose three methods: DMP-DUN$_\text{(10-step)}$ with $\Delta t=10$, DMP-DUN$^+_\text{(4-step)}$ with $\Delta t=25$, and DMP-DUN$^+_\text{(2-step)}$ with $\Delta t=50$. The FLOPs is calculated using an input image of size $256 \times 256$.

We trained on 118K images from COCO2017\cite{cocodataset} and validated on 100 images from BSDS500\cite{arbelaez2010contour}. Images are converted to YCbCr space, using Y-channel for training, validation and testing. Each training epoch crops 118K random $64\times 64$ patches from COCO2017 with randomly horizontal flipping. We used Adam\cite{kingma2014adam} as the optimizer with $(\beta_1, \beta_2)=(0.9, 0.999)$ and the default batch size is 8. The model is trained for 20 epochs with an initial learning rate of $1\times 10^{-4}$, which is decreased to $1\times 10^{-6}$ using the cosine annealing strategy\cite{Chen2022FSOINETFO, Loshchilov2016SGDRSG}. The model is implemented using PyTorch and trained on a single NVIDIA RTX3090 GPU.

\subsection{Reconstruction Quality Evaluation}
The reconstruction quality evaluation in this sub-section is divided into objective metrics comparison (\cref{table:main_ex_urban100} and \cref{table:main_ex_idm}) and subjective visual effects comparison (\cref{fig:reconstruction_compare}).

We utilize PSNR/SSIM as objective metrics to compare with several SOTA methods on Set11 \cite{kulkarni2016reconnet} and Urban100 \cite{Dong2018DenoisingPD}, as shown in \cref{table:main_ex_urban100}. It can be seen that our proposed DMP-DUN$^+_\text{(4-step)}$ achieves average PSNR values higher than PRL-RND$^+$\cite{DBLP:journals/ijcv/ChenSXZ23} by 1.39dB and 2.15dB on Set11 and Urban100, respectively, with fewer FLOPs by 258.8G. Additionally, our proposed DMP-DUN$^+_\text{(2-step)}$ improves the PSNR by 1.05dB and 1.49dB on Set11 and Urban100, respectively, while maintaining similar FLOPs to CPP-Net\cite{guoCPPNet2024}.

Additionally, we compare our method with the Invertible Diffusion Model (IDM)\cite{DBLP:journals/corr/abs-2403-17006}, which also employs end-to-end training of pre-trained diffusion models. Since IDM used $256\times 256$ crop size, we retrained DMP-DUN$^+_\text{(4-step)}$ under identical settings, as shown in \cref{table:main_ex_idm}. Our method achieves 1.23dB higher average PSNR than IDM. As IDM has demonstrated at least 5.55dB PSNR superiority over other diffusion-based methods\cite{DBLP:conf/iclr/SongVMK23, DBLP:conf/iclr/ChungKMKY23, DBLP:conf/cvpr/FeiLPZYLZ023, DBLP:conf/nips/RoutRDCDS23, DBLP:journals/pami/SahariaHCSFN23}, our approach logically surpasses these methods in reconstruction quality.

The subjective visual effects in \cref{fig:reconstruction_compare} show that our method is able to more realistically reflect the textures on the buildings without any blurring or confusing phenomena.

Moreover, since most of the methods compared in \cref{table:main_ex_urban100} did not use pre-trained models, to ensure the fairness of experiment, we compared the DMP-DUN$^+$ without using pre-trained weights with several methods\cite{ye2021csformer, DBLP:journals/ijcv/ChenSXZ23, guoCPPNet2024}, as shown in \cref{table:pretrained_ex}. It can be seen that average PSNR of DMP-DUN$^+_\text{(4-step)}$ without using pre-trained weights decreased by 0.42dB, but it still performed better than other state-of-the-art methods. This not only demonstrates that the powerful prior knowledge contained in pre-trained models can effectively improve reconstruction quality, but also shows that our algorithm is reasonable and robust, allowing it to perform well even without using pre-trained prior knowledge.

\subsection{Effect of Resblock for Decoupling Errors}
In our proposed DMP-DUN, we utilize ResBlock to achieve the acquisition of the AMP Onsager term, thereby ensuring that the difference between the output image and the true image (\ie the contained noise) follows a Gaussian distribution. \cref{fig:qq_plot} shows the Q-Q plots for the noise in output images of each module in DMP step of DMP-DUN$_\text{(10-step)}$ at different $ t $ during the Monarch image reconstruction process on the Set11 dataset. It can be observed from the figure that the noise in the images output by the ResBlock module basically follows a Gaussian distribution, confirming that it indeed can decouple the error from the image output by the preceding Gradient Descent module.

\subsection{Ablation Study}
We conducted an ablation study on the three modules of DMP step in DMP-DUN$^+$, as shown in \cref{table:abl_ex}. Cases (a), (b), and (c) demonstrate that all three modules in DMP significantly impact the reconstruction quality. Moreover, \cref{table:iterative_steps} shows the effect of different iterative steps on the reconstruction quality in DMP-DUN.

\subsection{Complexity study}

\begin{table}[t]
	\begin{adjustbox}{width=\linewidth,center}
		\begin{tabular}{@{}lccccc@{}}
			\toprule
			\multicolumn{1}{c}{\multirow{2}{*}{Metrics}} & \multicolumn{2}{c}{DDNM$^+_\text{(1000-step)}$\cite{wang2022zero}} & \multicolumn{3}{c}{DMP-DUN$^+_\text{(4-step)}$ (ours)} \\ \cmidrule(l){2-3}\cmidrule(l){4-6}  
			\multicolumn{1}{c}{} & 1 step & 1000 steps & 1 step & Side-ResBlcoks & 4 steps \\ \midrule
			FLOPs (G) & 67.0 & 67039.1 & 73.2 & 10.6 & 303.4 \\
			Params. (M) & 552.8 & 552.8 & 552.9 & 0.2 & 553.4 \\ \midrule
			PSNR (dB) & \multicolumn{2}{c}{32.34} & \multicolumn{3}{c}{24.07} \\ \bottomrule
		\end{tabular}
	\end{adjustbox}
	\caption{Comparison of DDNM and DMP-DUN$^+$ on average PSNR of Set11 and Urban100}
	\label{table:complex_ex_1}
\end{table}

\begin{table}[t]
	\centering
	\begin{adjustbox}{width=0.85\linewidth,center}
		\begin{tabular}{@{}ccccc@{}}
			\toprule
			\multirow{2}{*}{Metrics} & \multicolumn{2}{c}{Monte-Carlo SURE\cite{DBLP:journals/tip/RamaniBU08}} & \multicolumn{2}{c}{ResBlock (ours)} \\ \cmidrule(l){2-3} \cmidrule(l){4-5} 
			& 1 step & 4 steps & 1 step & 4 steps \\ \midrule
			\multicolumn{1}{l}{FLOPs (G)} & 134.1 & 536.4 & 6.1 & 24.4 \\
			\multicolumn{1}{l}{Params. (M)} & 0 & 0 & 0.09 & 0.37 \\ \bottomrule
		\end{tabular}
	\end{adjustbox}
	\caption{Comparison of using ResBlock to replace Monte-Carlo SURE for computing the Onsager term in DMP-DUN.}
	\label{table:complex_ex_2}
\end{table}

The complexity of DMP-DUN$^+$ compared to DDNM$^+$\cite{wang2022zero}, which is also based on the diffusion model approach, is illustrated in \cref{table:complex_ex_1}. Our method increases FLOPs by 6.2G per step but reduces the number of steps from 1000 to 4, while improving the reconstruction quality.

The complexity of using ResBlock as a replacement for Monte-Carlo SURE\cite{DBLP:journals/tip/RamaniBU08} is shown in \cref{table:complex_ex_2}. It can be seen that the use of ResBlock for approximation can save a total of approximately 512.0G FLOPs.

\section{Conclusion}
In this work, we design an iterative optimization algorithm named DMP, which leverages the powerful prior knowledge of pre-trained diffusion models while maintaining the manifold-constrained states according to its state evolution. We then unfold DMP into networks named DMP-DUN and DMP-DUN$^+$, which utilize lightweight neural networks to approximate Gaussian filters and Monte-Carlo SURE through end-to-end learning. Experimental results demonstrate the superiority of our proposed DMP-DUN and DMP-DUN$^+$ in terms of reconstruction quality and FLOPs compared to other state-of-the-art methods.

Since our method is derived through formula derivation and does not rely on specific diffusion model structures and pre-trained models, our method can further develop with the current research on diffusion models in the future.

\section*{Acknowledgments}

This work was supported by the Joint Fund of the Ministry of Education for Equipment Preresearch under Grants 8091B022121 and National Science and Technology Innovation 2030 (STI2030) Major Projects under Grants 2022ZD0205005.


{
	\small
	\bibliographystyle{ieeenat_fullname}
	\bibliography{main}
}

\end{document}